\documentclass{article}

\usepackage{xcolor}

\usepackage[preprint]{corl_2026} 

\usepackage{graphicx}
\usepackage{booktabs}
\usepackage{pifont}
\usepackage{multirow}

\newcommand{\cmark}{\ding{51}}
\newcommand{\xmark}{\ding{55}}

\title{\milo, a Fully Autonomous Indoor/Outdoor \\ Robotic Guide Dog}

\author{
  Florian Golemo\thanks{Equal contribution}\\
  Mila - Quebec AI Institute\\
  Montreal, Canada\\
  \texttt{fgolemo@gmail.com} \\
  \And
  Joanna Wolski\footnotemark[1]\\
  Mila - Quebec AI Institute\\
  Montreal, Canada\\
  \texttt{joannawolski@gmail.com} \\
  \And
  Joel Ruben Antony Moniz\\
  Mila - Quebec AI Institute, Polytechnique\\
  Montreal, Canada\\
  \And
  Christopher Pal\\
  Polytechnique, Mila - Quebec AI Institute\\
  Montreal, Canada\\
}

\usepackage[dvipsnames]{xcolor}
\usepackage{enumitem}
\usepackage{wrapfig}
\usepackage[title]{appendix}
\usepackage{amsmath}

\newcommand{\green}[1]{\textcolor{OliveGreen}{#1}} 
\newcommand{\red}[1]{\textcolor{red}{#1}} 
\newcommand{\orang}[1]{\textcolor{BurntOrange}{#1}} 
\newcommand{\milo}{Mi\textit{l}o} 
\begin{document}
\maketitle

\vspace{-25pt}
\begin{figure}[h]
\centering
\includegraphics[width=0.7\textwidth]{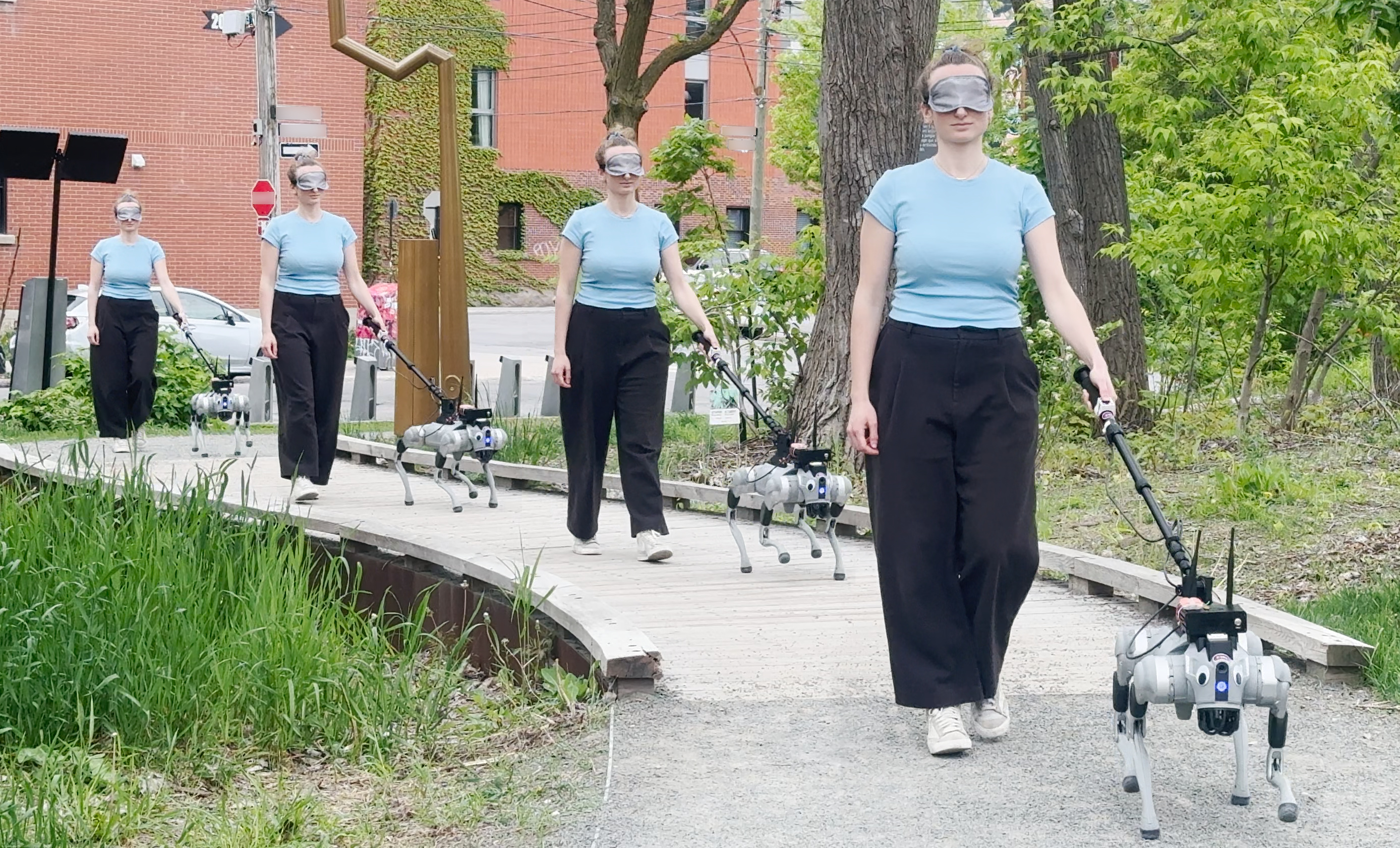}
\caption{\textbf{\milo} autonomously guides a handler along indoor and outdoor paths while avoiding obstacles and oncoming pedestrians, adapting to the handler's position, desired walking speed, and turn commands. The project is released as open source. }
\label{fig:cover}
\end{figure}
\vspace{-5pt}

\begin{abstract}

Many Blind and Low-Vision (BLV) people rely on guide dogs for moment-to-moment navigation, such as staying on path and avoiding obstacles and pedestrians. 
However, guide dogs are expensive to acquire and maintain (approximately \$50k USD plus ongoing costs), often involve long waiting lists, and have relatively short life expectancies. 
While robot guide dogs offer a promising alternative, existing approaches exploring this idea suffer from several drawbacks:
They often lack the autonomy required for real-world deployment, relying on prior 3D scans of the environment, external computation, or limited awareness of the handler.
In this work, we present \milo, the first open-source, low-cost (approximately \$2k USD) robotic guide dog platform capable of fulfilling the basic collaborative navigation role expected of a guide dog.  
\milo\ is fully autonomous, requiring no \textit{a priori} knowledge of the environment, completely self-contained with all computation performed onboard, and suitable for both indoor and outdoor navigation while avoiding obstacles and pedestrians.
Our system consists of a modified Unitree Go2 robot (equipped with onboard compute, sensors, and a handle), a perception stack combining voxel mapping with floor, obstacle, and pedestrian detection, and a navigation stack based on an obstacle-avoidance policy trained in a custom bird's-eye-view simulator.
We evaluate \milo\ in real indoor and outdoor obstacle courses and compare it against a costmap-based baseline, demonstrating smoother navigation and fewer handler collisions.
To maximize accessibility for BLV users, we release both the robot hardware instructions and the complete software stack as open source\footnote{The Milo project website is available here: \url{https://fgolemo.github.io/milo}}.

\end{abstract}



\section{Introduction}
\label{sec:intro}	

According to recent global estimates, approximately 43 million people worldwide are blind and more than 295 million live with moderate-to-severe vision impairment~\cite{bourne_trends_2021}. 
Many Blind and Low Vision (BLV) individuals rely on white canes, electronic navigation aids, or trained guide dogs for safe and independent mobility. 
However, just breeding, raising, and training a guide dog can cost more than \$50{,}000 per animal~\cite{guide_dog_foundation}. 
Additionally, prospective handlers often face long waiting lists before being matched with a dog. 
These challenges motivate the development of robotic guide dog systems capable of providing accessible and scalable mobility assistance. \\
A robotic guide dog must safely navigate complex environments by identifying walkable paths, avoiding obstacles and pedestrians, and maintaining comfortable motion for the handler. 
Several recent works have explored robotic guide dog navigation~\cite{sorokin2022learning, cai2024navigating, hwang2022system, hwang_guidenav_2025, hayamizu_woofs_2026}. 
However, many existing systems rely on prior maps~\cite{cai2024navigating}, teleoperation ~\cite{hayamizu_woofs_2026}, or external computation~\cite{hwang2022system}, limiting their autonomy in real-world deployment. 
Another limitation is that existing approaches often treat the robot as an independent agent, even though guide dog navigation is inherently collaborative. 
In practice, the handler provides high-level navigation intent while the guide dog performs local path following and obstacle avoidance~\cite{hwang_towards_2024, hwang2022system}. 
A robotic guide dog should therefore adapt its behavior to the handler rather than operate in isolation.\\
To address these challenges, we propose a robotic guide dog navigation system for previously unseen indoor and outdoor environments that is capable of operating autonomously on-device. 
Our approach combines walkable-path segmentation, obstacle avoidance, and reinforcement-learning-based navigation with explicit handler awareness, enabling operation without prior mapping or external infrastructure.
We developed a GPU-accelerated bird's-eye view (BEV) simulator, implemented in Taichi \cite{hu2019taichi}, that simulates a robot guiding a handler along indoor/outdoor sidewalks while avoiding obstacles and pedestrians.
To make the system robust and responsive to various handler needs, the navigation policy is trained via Reinforcement Learning (RL) using diverse handler positions.
At run time, onboard RGB images, LiDAR scans, odometry, and handle magnetic encoder measurements are acquired and processed into a human-interpretable BEV representation that matches the simulator observations.
The policy trained in simulation is deployed zero-shot on a modified quadruped robot.
The robot is equipped with a handle mounted through a two-degree-of-freedom passive joint that measures the handler's relative position. 
A directional (D)-pad on the handle allows the handler to adjust the walking speed and issue turn commands.
We evaluate our system in previously unseen indoor and outdoor environments on path following, static obstacle avoidance, and pedestrian avoidance. 
Compared with a costmap-based baseline, which follows paths but does not account for changes in the handler's position and suffers from blind spots, our BEV-based RL policy retains a geometric memory of the environment and protects the handler from collisions.\\
The main contributions of this work are:
\begin{enumerate}[nosep]
\item A fully onboard robotic guide dog navigation system capable of operating in previously unseen indoor and outdoor environments without prior mapping, consisting of custom robot hardware for dynamic handler motion and a perception/BEV-mapping system.
\item An indoor/outdoor BEV simulation environment for efficiently training and evaluating reinforcement learning policies for path-following and smooth obstacle avoidance policies.
\item An open-source release\footnotemark[2] of robot hardware design and the codebase including the simulation environment, policy training, onboard perception stack, and pretrained policies.
\end{enumerate}


\section{Related Work}
\label{sec:related}

\begin{table*}[tb]
\centering
\caption{Comparison with previous robotic guide navigation systems.}
\label{tab:comparison}
\resizebox{\textwidth}{!}{
\begin{tabular}{lcccccc}
\toprule
Method & 
\begin{tabular}[c]{@{}c@{}}Indoor \& Outdoor\\ Navigation\end{tabular} & 
\begin{tabular}[c]{@{}c@{}}Unknown\\ Environment\end{tabular} & 
\begin{tabular}[c]{@{}c@{}}Fully\\ Onboard\end{tabular} &
\begin{tabular}[c]{@{}c@{}}Takes\\ Handler\\ Commands\end{tabular} & 
\begin{tabular}[c]{@{}c@{}}Dynamic\\ Handler\\ Modeling\end{tabular} &
\begin{tabular}[c]{@{}c@{}}Open\\ Source\end{tabular} \\
\midrule
\citet{sorokin2022learning} & \orang{Outdoor only} & \green{\cmark} & \green{\cmark} & \red{\xmark} & \red{\xmark} & \red{\xmark} \\
\citet{cai2024navigating}, ``RDog'' & \green{\cmark} & \red{\xmark} & \green{\cmark} & \green{\cmark} & \orang{Partial} & \red{\xmark} \\
\citet{hwang2022system}, ``Summer'' &  \orang{Indoor only} & \green{\cmark} & \red{\xmark} & \green{\cmark} & \orang{Static} & \red{\xmark} \\
\citet{hwang_guidenav_2025}, ``GuideNav'' & \orang{Outdoor only} & \red{\xmark} & \green{\cmark} & \red{\xmark} & \red{\xmark} & \red{\xmark} \\
Ours, ``\milo'' & \green{\cmark} & \green{\cmark} & \green{\cmark} & \green{\cmark} & \green{\cmark} & \green{\cmark} \\
\bottomrule
\end{tabular}
}
\end{table*}

\textbf{Hardware Autonomy.} Hardware autonomy is a fundamental requirement for assistive guide robots operating in real-world environments. 
However, several existing systems still offload perception or planning pipelines to external laptops or remote computation \cite{hwang2022system}. 
Recent work has improved onboard integration by embedding compact high-performance computing hardware directly onto the robot platform \cite{hwang_guidenav_2025}\cite{sorokin2022learning}, enabling fully onboard navigation without external resources. 
Similarly, our framework performs the entire pipeline onboard the robot.\\
\textbf{Perception and Navigation.} Previous work has demonstrated autonomous navigation in either indoor \cite{hwang2022system}\cite{chen_exploration_2025},  or outdoor environments \cite{sorokin2022learning}\cite{viteri_autonomous_2024}. 
Indoor systems often rely primarily on LiDAR data and assume all traversable space is walkable \cite{chen_exploration_2025}, an assumption that does not generalize well outdoors, where sidewalks, roads, and other terrain types must be distinguished. 
Consequently, outdoor navigation approaches commonly integrate semantic segmentation modules to identify walkable pedestrian regions \cite{sorokin2022learning}. 
However, many of these systems depend heavily on GPS localization \cite{sorokin2022learning} \cite{viteri_autonomous_2024}, limiting deployment in indoor environments. 
In contrast, our segmentation framework enables indoor and outdoor navigation without GPS.\\
Generalization to unknown environments also remains challenging. 
Many systems rely on prior map information from services such as Google Maps \cite{viteri_autonomous_2024} or require building custom 3D maps of the environment beforehand \cite{cai2024navigating}, while others learn to follow previously teleoperated routes \cite{hwang_guidenav_2025}. 
To improve navigation and obstacle avoidance in such environments, several approaches construct bird’s-eye-view (BEV) representations of the surroundings \cite{sorokin2022learning,cai2024navigating}, enabling the robot to maintain a spatial memory of nearby obstacles and free space. 
Multi-modal perception systems combining LiDAR and RGB sensing \cite{cai2024navigating,sorokin2022learning} generally provide more robust obstacle detection and mapping than camera-only approaches \cite{viteri_autonomous_2024,hwang_guidenav_2025}, particularly by reducing close-range blind spots and improving depth estimation. 
Unlike these systems, our framework does not require prior knowledge of the environment and incrementally builds a BEV map of the environment using onboard LiDAR and RGB sensors.\\
For obstacle avoidance and navigation planning, classical approaches typically rely on ground-plane costmaps to compute collision-free trajectories \cite{hwang2022system}, while more recent methods train reinforcement learning policies in simulated BEV environments \cite{sorokin2022learning}. 
These recent learning-based approaches demonstrate promising generalization to unknown environments, but training often remains computationally expensive, requiring approximately 30 hours \cite{sorokin2022learning}. 
Our approach instead trains a navigation policy in approximately 10 minutes.\\
\textbf{Handler Modeling.} Guide dog mobility relies on continuous commands from the handler \cite{hwang_towards_2024}. 
To reproduce this interaction in assistive robotics, prior work has explored rigid handle interfaces \cite{kim2023transforming} \cite{hwang2022system} \cite{cai2024navigating}, and joystick-based controls \cite{hwang2022system} \cite{cai2024navigating}, allowing the handler to communicate directional intent and walking preferences.
However, most existing guide robot systems oversimplify the handler’s physical presence during navigation by assuming a fixed geometric offset relative to the robot body \cite{hwang2022system}. 
Consequently, obstacle avoidance policies primarily plan trajectories for the robot itself, potentially generating paths that are safe for the robot but unsafe for the handler.
Our work addresses this limitation by integrating joystick-based handler commands through a rigid guidance handle while explicitly modeling the handler’s dynamic spatial footprint within the obstacle avoidance and navigation loop.


\section{Method}
\label{sec:method}

\begin{wrapfigure}[20]{r}{0.5\textwidth}
\vspace{-50pt}
  \begin{center}
    \includegraphics[width=0.5\textwidth]{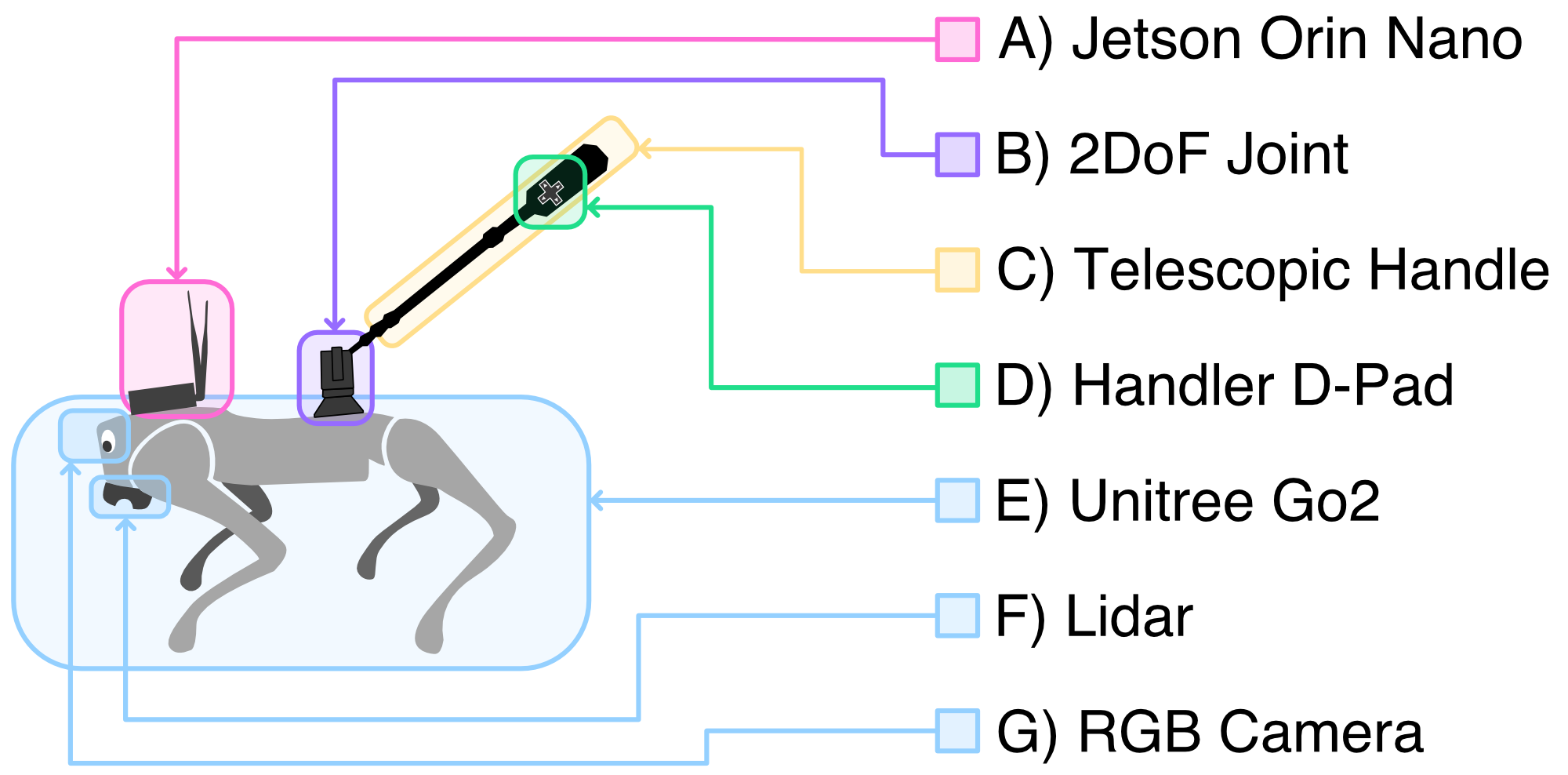}
  \end{center}
\caption{\textbf{Hardware Overview.} (A) The Nvidia Jetson Orin Nano runs navigation and perception systems; (B) the 2 degree-of-freedom joint is equipped with magnetic encoders that sense the handler's position with respect to the dog; (C) the telescopic handle is adjustable to cater to different handler heights and preferences; (D) with the 5-way directional pad, the handler can start/stop the robot, indicate faster/slower walk speed, and desire to turn at the next opportunity; (E) the whole system is based on a Unitree Go2 Air/Pro robot dog, utilizing its (F) LiDAR and (G) front-facing RGB camera.}
  \label{fig:system}
\end{wrapfigure}

\begin{figure}[t]
\centering
\includegraphics[width=1.0\textwidth]{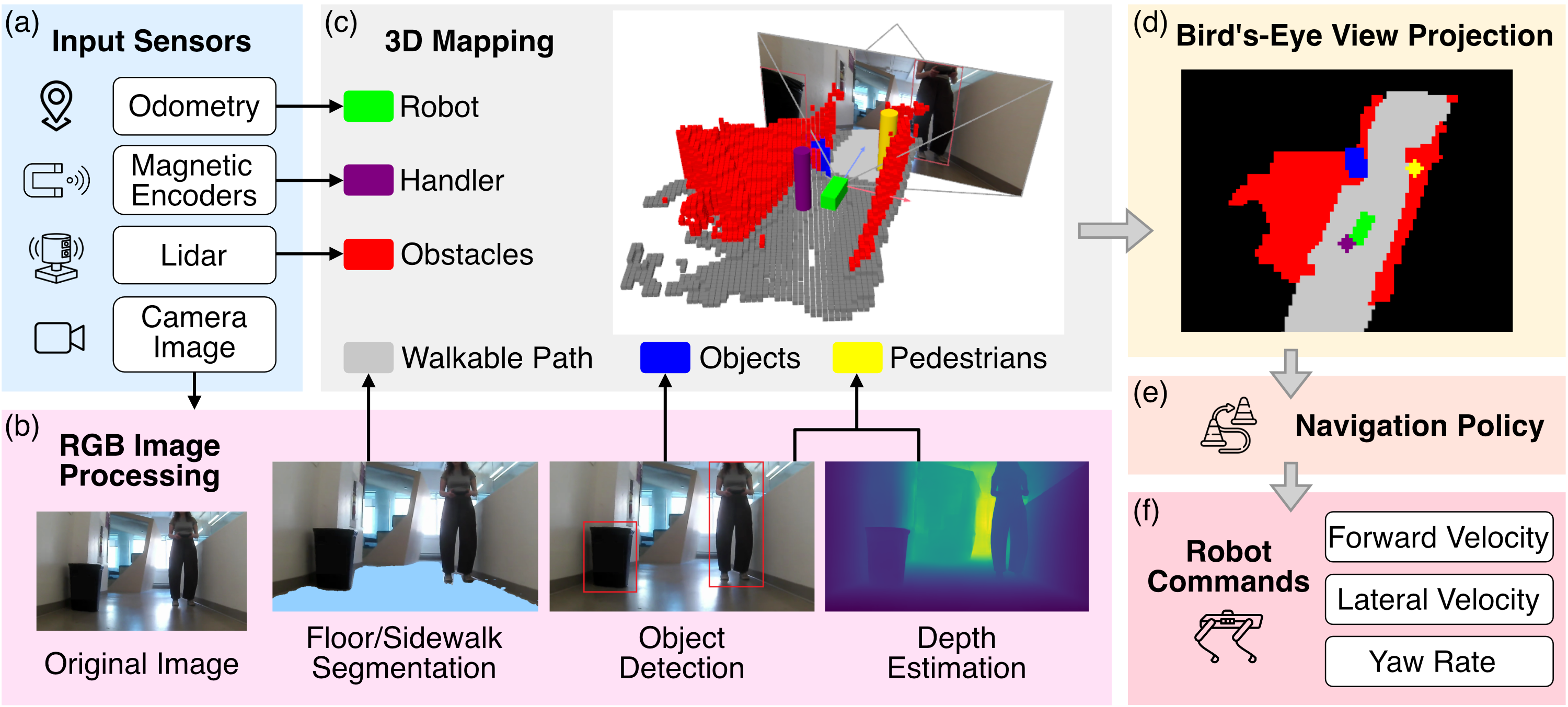} 
\caption{\textbf{Method Overview.} (a) RGB images, LiDAR scans, odometry, and handle encoder measurements are acquired. (b) RGB images are processed to segment the walkable path and detect objects and pedestrians, while depth estimation updates pedestrian positions. (c) The segmented path, detected objects and pedestrians, the robot and handler states, and LiDAR data are fused into a voxel-based 3D map. (d) The voxel map is converted into a bird's-eye-view (BEV) representation. (e) The BEV map is provided to the trained navigation policy, which predicts the robot's linear velocity and yaw rate commands. (f) The predicted commands are filtered by a local obstacle avoidance safety module before being sent to the robot.}
\label{fig:method}
\end{figure}

Rather than learning navigation end-to-end as a black-box policy, our approach constructs a human-interpretable bird's-eye-view (BEV) representation of the environment, which serves as the policy input. 
Compared with a dense 3D map, the BEV representation is more memory-efficient while remaining suitable for downstream accessibility features, such as audio cues and natural-language scene descriptions (e.g., "obstacle ahead in 2 meters at 10 o'clock").
The navigation policy is trained offline in a GPU-accelerated 2D BEV simulator to follow randomly generated paths while avoiding obstacles and pedestrians.
At run time, the onboard perception and control pipeline, illustrated in Fig.\ref{fig:method}, acquires RGB images, LiDAR scans, odometry, and handle encoder measurements.
RGB images are processed with NanoSAM \cite{mobile_sam, nvidia_nanosam} to segment the walkable path and with YOLO \cite{yolo26_ultralytics} to detect objects and pedestrians.
Pedestrian positions are estimated using Metric Depth Anything V2 \cite{depth_anything_v2}, providing faster updates than relying solely on the LiDAR map.
A 3D voxel-based map is constructed from temporally accumulated LiDAR scans, into which the segmented path, object detections, pedestrian positions, and the robot and handler states are fused.
The map is then converted into a simplified BEV representation and serves as input to the trained navigation policy.
Its predicted velocities and yaw rate commands are filtered by a local obstacle avoidance safety filter (Appendix App.\ref{app:safety}) before being scaled according to the handler's preferred walking speed and sent to the robot.


\subsection{Hardware}
\label{sec:method-hardware}

While the proposed navigation and perception framework does not depend on a specific robot platform, it was designed with low-cost quadruped robots in mind, such as the Unitree Go2 Air/Pro robot dog\footnote{\url{https://www.unitree.com/go2}}. 
We chose the Go2 for its affordability, front-facing camera and LiDAR, and wide availability.
The robot does not come equipped with compute hardware that can run machine learning inference, which required us to add a compute module. The Nvidia Jetson Orin Nano\footnote{\url{https://www.nvidia.com/en-us/autonomous-machines/embedded-systems/jetson-orin/nano-super-developer-kit/}} struck a good balance between price, size, and capabilities. 
In our experiments, it ran our complete navigation pipeline at 5 Hz, including voxel mapping, the segmentation and object detection networks, and the CNN-based policy.\\
Guide dogs are typically held in the handler's left hand, allowing the right hand to operate a white cane \cite{hwang_towards_2024}. However, some left-handed handlers might prefer to switch sides.
Flexible handler positioning enables the robotic guide dog to adapt its relative configuration when traversing narrow spaces, for example by temporarily moving ahead of the handler to avoid nearby obstacles.
Supporting this flexibility required a custom physical interface; instead of a rigid harness, we use a telescopic handle that accommodates different handler-robot distance preferences and a custom 3D-printed 2 degrees-of-freedom (DoF) joint to securely attach the handle to the robot.
The joint is equipped with magnetic encoder sensors to measure the azimuth and elevation angles of the handle in real-time.
A five-way directional pad (D-pad) mounted at the end of the handle provides user commands, including pause/resume (center button), desired velocity adjustments (forward/backward), and turn requests (left/right).
See Fig.\ref{fig:system} for a diagram of the complete hardware configuration.

\subsection{Navigation}
\label{sec:method-nav}

In order to safely learn how to follow sidewalks and navigate indoor spaces, we built a sidewalk simulator similar to \citet{sorokin2022learning}\footnote{Sadly, their  project page is no longer available, so we were not able to use their code if it was released.}.
The simulator was implemented in Taichi \cite{hu2019taichi} to support GPU acceleration across devices, comparable to massively-parallel locomotion learning framework from \citet{rudin2021learning}.
During training, the simulator generates random environments consisting of paths, intersections, walls, obstacles, and pedestrians surrounding the robot–handler pair (see Fig.\ref{fig:frame-sim} for an example frame). 
The robot is perpetually moved forward in the simulator and can modulate its forward, lateral, and yaw speeds as an action.
It is incentivized to maintain a moderate forward velocity while avoiding obstacles. 
For more implementation details of the simulator, please see appendix section App.\ref{app:sim}, ``Simulator Implementation.'' 
In order to train a navigation policy in simulation, we express the problem as a Markov decision process with the following properties.

\textbf{Observation Space.} The observations are bird's-eye view RGB images with a history of 2 additional previous observations. 
Please see Fig.\ref{fig:frame-sim} for an annotated example.

\textbf{Action Space.} The actions are forward and lateral velocity as well as yaw command. 
The forward speed component of the action is summed with the default forward velocity in simulation, while the lateral and yaw speed action components are applied to the robot's body verbatim.

\textbf{Rewards.} The reward function consists of 14 terms: eight binary collision penalties defined by the $[dog,handler] \times [nonroad,wall,obstacle,pedestrian]$ collision matrix, where each term is activated when the corresponding collision occurs, a Gaussian velocity-tracking term $r_{vel}$ to encourage moving forward, two quadratic lateral movement/yaw tracking terms $r_{lat}$ and $r_{yaw}$ to encourage minimal lateral/yaw movement, a jerk penalty term to reduce sudden motion $r_{jerk}$, and two rewards for keeping circular safety zones around the robot and handler clear, an inner zone $r_{inner\_danger}$ and an outer zone $r_{outer\_danger}$.
Please find details about the implementation of these terms and their respective coefficients in the appendix App.\ref{app:rl}, ``Reinforcement Learning''.

\textbf{Policy Learning.} Policy learning is implemented through GPU-optimized Proximal Policy Optimization (PPO), following \citet{rudin2021learning}.

\subsection{Perception}
\label{sec:method-perc}

The perception pipeline is designed to provide a real-time spatial representation of the robot, the handler, and their surrounding environment. 
To achieve this, the system combines data from onboard LiDAR, RGB camera, and magnetic encoders to construct a unified bird’s-eye view (BEV) representation of the scene. 
Images captured by the onboard camera are first rectified using intrinsic calibration parameters to remove lens distortion and fisheye effects.

\textbf{BEV Representation.} Inspired by BEVFusion \cite{liang2022bevfusion}, we construct a top-down semantic representation of the environment centered on the robot. 
The BEV map encodes walkable regions, obstacles, detected objects and pedestrians, as well as the robot and handler positions. 
Object detections provide additional semantic context beyond obstacle occupancy, while explicitly representing pedestrians enables smoother interactions with moving entities.
Fig.\ref{fig:method} illustrates the different components of the generated BEV representation.
The resulting BEV images are used as observations for the navigation policy described in Section~\ref{sec:method-nav}.

\textbf{LiDAR Processing.} We use the LiDAR map produced by the robot’s onboard localization and mapping system and stream it in real time through the robot WebRTC interface. 
The resulting point cloud is motion-corrected to account for robot movement during LiDAR acquisition.
The map is represented as a voxel-based 3D structure and downsampled into a lower-resolution occupancy grid before projection onto a 2D BEV plane. 
Voxels below a height threshold are removed to filter ground points while remaining robust to uneven terrain.  
Similarly, voxels above a maximum height threshold are discarded to ignore elevated structures such as ceilings or tree canopies that the robot and handler can safely pass under.
The resulting occupancy grid represents surrounding obstacles and non-traversable regions and is continuously updated online as the robot moves through the environment.

\textbf{Walkable Path Segmentation.}  Walkable path segmentation (e.g., indoor floors or outdoor sidewalks) is computed directly from RGB images. 
Our approach uses NVIDIA NanoSAM\footnote{\url{https://www.jetson-ai-lab.com/archive/vit/tutorial_nanosam.html}}, a distillation model optimized for Jetson devices based on MobileSAM \cite{mobile_sam}. 
NanoSAM generates segmentation masks from prompt points placed near the lower center of the image, assuming the robot is initially positioned on and facing a walkable path. The resulting segmentation mask is projected into the BEV representation.

\textbf{Object and Pedestrian Detection.} Objects and pedestrians are detected using Ultralytics YOLO26n \cite{yolo26_ultralytics} on rectified camera frames. 
Detections are associated with LiDAR points through geometric projection between the camera and LiDAR coordinate frames, then projected onto the 2D BEV map. 
Pedestrians are represented separately from generic obstacles to encourage smoother interactions with moving entities.
To improve long-range pedestrian detection beyond the LiDAR occupancy map, we use monocular depth estimation with Depth Anything \cite{depth_anything_v2}. 
Depth predictions are calibrated with LiDAR measurements and are used to estimate pedestrian distance (see Appendix App.\ref{app:depth}, ``Pedestrian Depth Estimation'').

\textbf{Magnetic Encoder Processing.} The handler's position relative to the robot is estimated using two magnetic encoders mounted at the base of the guiding handle. 
The handle length and measured azimuth and elevation are used to compute the relative 3D position $(x, y, z)$ of the handler in the robot frame. 
This position is represented as a circular region in the BEV map, allowing the navigation policy to account for the handler during navigation.


\section{Experiments}
\label{sec:exp}

\subsection{Experimental Setup}
\label{sec:exp-setup}

\begin{wrapfigure}[22]{r}{0.45\textwidth}
\vspace{-5pt}
  \begin{center}
    \includegraphics[width=0.45\textwidth]{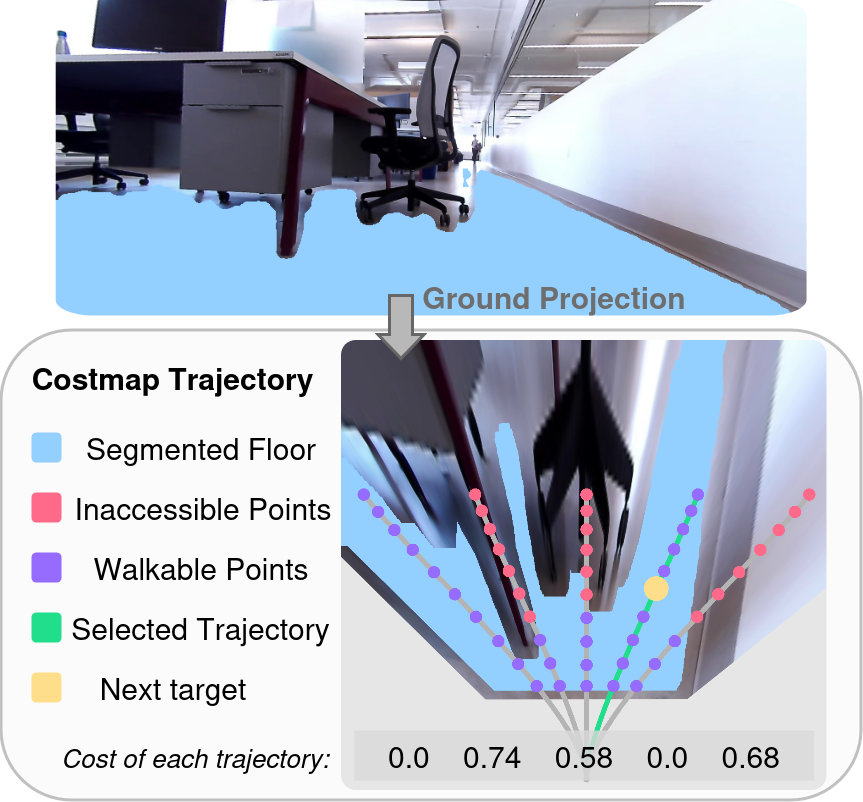}
  \end{center}
\caption{\textbf{Example of Costmap-Based Trajectory Selection.} The segmented camera frame is projected onto the ground plane to generate trajectories. The center-right trajectory is selected as it has the lowest cost; ties are resolved in favor of the rightmost trajectory.}
 \label{fig:costmap}
\end{wrapfigure}

\textbf{Costmap Baseline.} As a baseline for evaluating our learned navigation policy, we implemented a classic costmap-based controller similar to \citet{hwang2022system}.
We first project the previously segmented walkable-path mask onto the ground plane using a calibrated homography transformation. 
From this ground projection, we generate five candidate trajectories corresponding to \textit{extreme left}, \textit{left}, \textit{straight}, \textit{right}, and \textit{extreme right} motions. 
Each trajectory is modeled as a smooth Bézier curve to ensure forward and kinematically feasible motion.
To evaluate a trajectory, we sample points along the curve and verify whether they remain within the traversable region see Fig.\ref{fig:costmap}. 
A traversal cost is then assigned based on how far along the trajectory it remains traversable. 
Trajectories that leave the walkable region early receive a high cost, whereas fully traversable trajectories receive a cost of zero.
The trajectory with the minimum cost is selected for execution. 
In the case of equal costs, we prioritize straight trajectories over lateral alternatives and right-side trajectories over left-side ones. 
Finally, an intermediate goal point is extracted from the selected trajectory and transformed into world coordinates to provide the target position for the robot controller.

\textbf{Control Loop For Policy Rollout.} Once the policy is trained in simulation, the actor network is JIT-compiled for better performance. 
On the robot, 5 threaded loops are running at different frequencies, some governed by the publishing speed of that sensor, some limited by the hardware:
(1) The LiDAR is accumulated into a voxel grid of 10cm resolution at 5Hz. 
This thread updates the walls in the BEV frame and provides the 3D geometry to which the floor and object segmentations are linked.
(2) The odometry thread reads the robot's position and orientation as well as the magnetic encoders to update the robot and handler positions in the BEV frame at 18Hz.
(3) The camera thread receives images of $1280\times720$px which are rectified to approximately $900\times400$px.
SAM and Yolo process the rectified frame and update object- and floor voxels for the BEV frame. 
Due to hardware constraints, this thread runs at 4Hz.
(4) The policy extracts the latest BEV frame as a 200 × 200 × 3 image, stacks it with the two previous observations to form a 200 × 200 × 9 input, performs inference, and outputs a control command at 4 Hz (limited by the camera thread).
(5) The control thread runs at 10Hz, repeating the last command from the policy while waiting for the next command. 

\textbf{Additional Safety Filter.} In both cases, any action given by the policy or costmap is filtered through a local obstacle avoidance system that has direct access to lidar frames and robot odometry and serves as an additional means of thwarting collisions before the action is sent to the robot.
The system is explained in more detail in the appendix section App.\ref{app:safety}, ``Local Obstacle Safety Filter''.

\begin{figure}[t]
\centering
\includegraphics[width=0.9\textwidth]{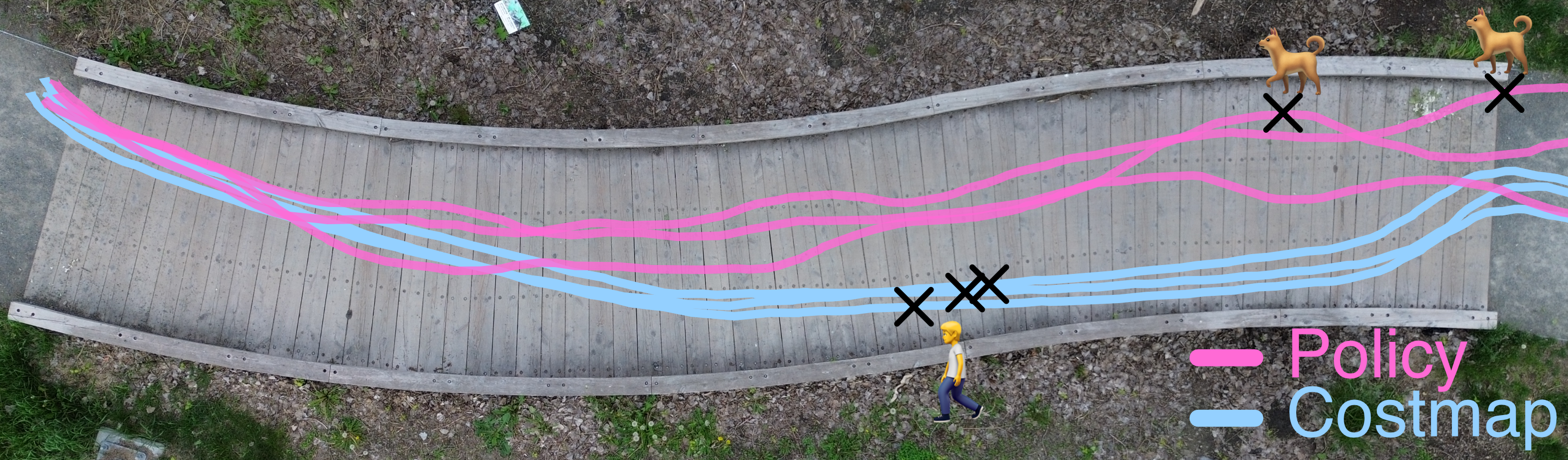}
\caption{\textbf{Experiment - Path Following.} We compared the costmap controller (cyan) at fixed speed to the policy (pink) at variable speed. Starting at the left, both approaches had to cross the bridge and were timed at the start and finish. The costmap led the handler out of bounds for several meters (start marked on the map with X). The policy incurred 2 robot collisions (marked with X) at higher speeds but otherwise stayed center on the path.}
\label{fig:exp-outdoor}
\end{figure}

\subsection{Path Following}
\label{sec:exp-straight}

To verify that our system can follow paths without walls, solely relying on floor segmentation, we picked an outdoor path on an S-shaped wooden bridge in a park area (Fig.\ref{fig:exp-outdoor}, same route as shown in Fig.\ref{fig:cover}).
The starting position and orientation were picked so that the robot would need to turn at least twice to traverse safely with the handler following on the right side.
We recorded 3 runs with the costmap controller with consistent speed and 3 runs of the policy with 3 different walking speeds to test how fast our system can traverse the course.

The costmap controller took an average of 21.3s $\pm$1s and the policy ranged from 13.7s at the highest speed to 22.3s at the lowest.
Since the costmap controller does not have any awareness of the handler's relative position, every run incurred a handler collision with the boundary of the course, and the robot followed a path unsafe for the handler for several meters. 
In the policy, we noticed 2 collisions of the robot at higher speeds for which we had to intervene and there were no handler collisions.

\begin{wrapfigure}[14]{r}{0.40\textwidth}
  \vspace{-42pt}
  \begin{center}
    \includegraphics[width=0.4\textwidth]{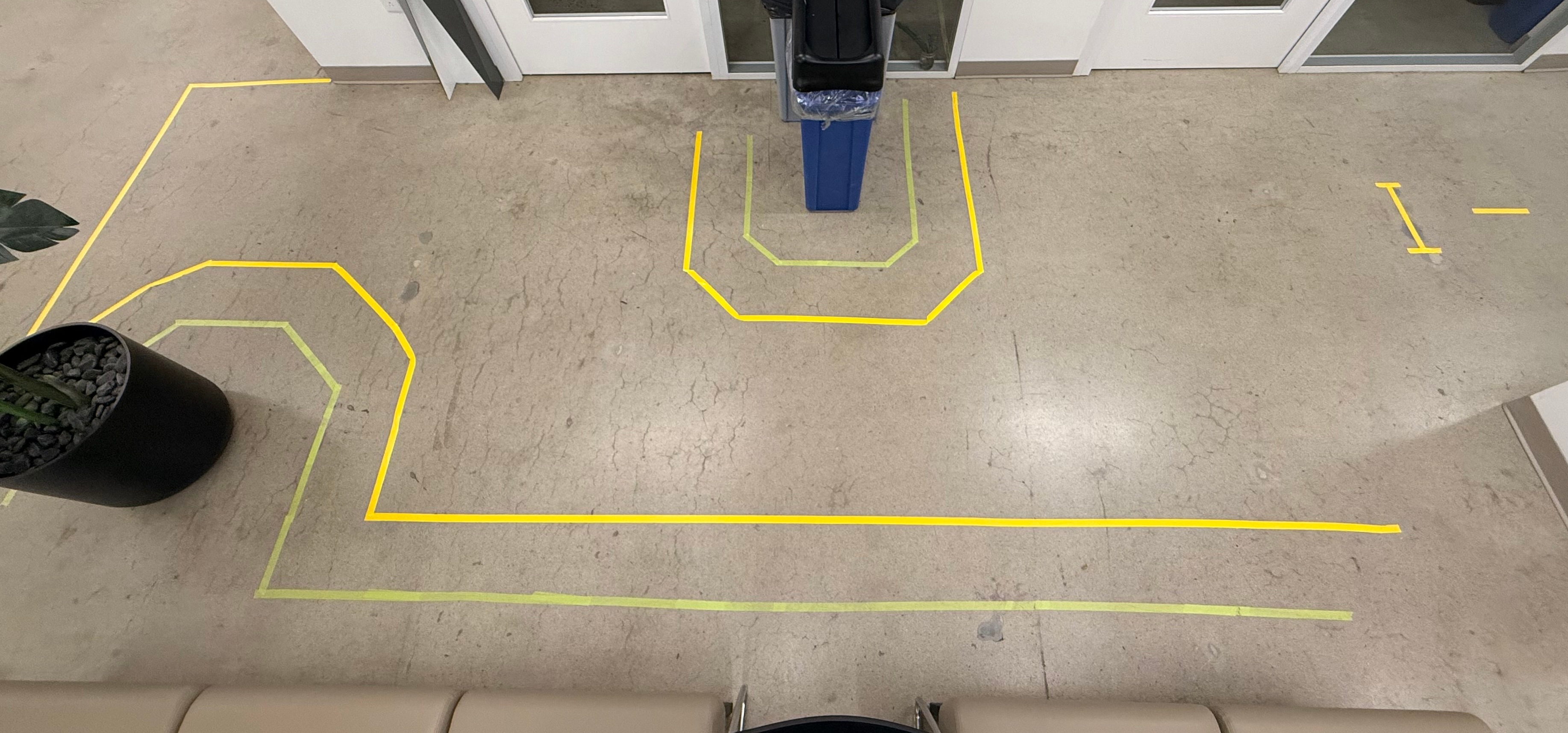}
  \end{center}
\caption{\textbf{Experiment - Obstacle Avoidance.} \milo\ and the handler started at the top right facing left and navigated around a rubbish bin and a planter. All costmap runs resulted in handler collisions with the bin, while the policy runs protected the handler, although two grazed the planter.}
  \label{fig:exp-obst}
\end{wrapfigure}




\subsection{Static Obstacle Avoidance}
\label{sec:exp-avoid}

We also tested turning around static obstacles, for which we set up a relatively tight turn for the robot \& handler unit.
We marked ``comfort'' areas on the floor around the obstacles with one line indicating 0.25m (dangerously close) and another line at 0.5m around obstacles.
Following the same experimental protocol as before, we recorded 3 rollouts with the costmap controller and 3 runs with the policy.
The completion time of the policy is slightly higher at 11.7s $\pm$2s compared to the costmap 10.5s $\pm$1s.
All costmap runs produced hard collisions of the handler with the first obstacle (rubbish bin) and 2/3 runs had the robot come dangerously close to the bin (under 0.25m).
The policy runs did not include any dangerous handler trajectories, but in all rollouts, the robot walked dangerously close to the second obstacle and touched it in 2/3 runs. 
As before, we observed the policy being better at shielding the handler while incurring some collisions between the robot and the environment.



\subsection{Pedestrian Avoidance}
\label{sec:exp-obst}

Our intention was to evaluate the comfort (minimum distance) and number of  handler and robot collisions in an obstacle course consisting of 2 collaborators walking towards the robot and handler in a corridor at a fixed pace, but in practice, we found little repeatability in this setup between experiments. 
Qualitatively across multiple indoor and outdoor runs, \milo\ was able to avoid oncoming pedestrians but struggled with pedestrians overtaking from behind. 
The latter scenarios sometimes led to side-stepping behaviors when the overtaking pedestrian first came in sight and until they were at least a meter past \milo.
We believe this behavior can be addressed by modulating the lateral and yaw response when pedestrians suddenly appear within a 1m radius.
We included samples of unplanned outdoor pedestrian avoidance in the supplementary video: \url{https://youtu.be/hCJE4ORoizo}\\


\section{Conclusion}
\label{sec:conclusion}

In this work, we presented \milo, a fully self-contained, low-cost robotic guide dog designed to assist Blind and Low-Vision (BLV) users during navigation. 
The proposed system operates in previously unseen indoor and outdoor environments without requiring prior maps, external infrastructure, or internet connectivity. 
By combining lightweight onboard perception with a reinforcement-learning-based navigation policy, \milo\ learns to follow walkable paths while avoiding static and dynamic obstacles, including pedestrians.
Our experiments demonstrate that \milo\ can reliably perform collaborative navigation in real-world environments. 
Compared to a classical costmap-based baseline, the learned policy reduced handler collisions and provided safer guidance for the user. 
Nevertheless, several limitations remain, as discussed below, motivating future work.
By releasing the complete hardware and software stack as open source, we hope to support future research on robotic guide dogs and contribute to the development of more accessible mobility assistance technologies for the BLV community.


\section{Limitations and Future Work}
\label{sec:limitations}

Our system has several limitations that motivate future work. First, the NanoSAM-based walkable-path segmentation assumes that the robot is already positioned on and facing a traversable surface, since the prompt points are placed near the lower center of the image. This limitation could be addressed by first using NanoOwl\footnote{\url{https://github.com/NVIDIA-AI-IOT/nanoowl}} to localize semantic concepts such as sidewalks or floors and provide a bounding-box prompt for NanoSAM. Second, the learned policy does not always balance handler safety, robot safety, and social navigation. While it generally provides safer trajectories for the handler than the costmap baseline in simple environments, it may reduce the robot's own clearance from obstacles. In addition, the robot tends to behave overly conservatively around pedestrians in crowded environments, despite people often approaching it out of curiosity. Future work could investigate socially-aware navigation policies that better optimizes these competing objectives. A further limitation is that the current ground-projection method assumes flat terrain, and handling uneven surfaces such as slopes and ramps remains future work. LiDAR map updates can also become a bottleneck at higher walking speeds. Accessing raw LiDAR data through the DDS interface could improve update rates, at the cost of increased processing complexity. Finally, the current framework is limited to sidewalk and floor navigation and cannot cross streets. Future work could extend the system to more complex urban environments by incorporating pedestrian crossing, traffic light, and vehicle detection, while also providing richer voice feedback to improve the handler's understanding of the surroundings, similar to the approach proposed by \citet{hayamizu_woofs_2026}.



\clearpage
\acknowledgments{The authors would like to thank Luis Lara and Amaury Wei for their continued advice and assistance with recording the robot experiments.
We'd also like to thank the Mila Ecosystem team for allowing us to put down tons of tape in the hallways to measure robot progress and calibrate different parts of the perception system.
Thanks to Kirsty Ellis for lending us a spare Nvidia Jetson Orin dev board.
We'd also like to thank the folks from DimOS for providing the inspiration to some of our perception stack (\url{https://github.com/dimensionalOS/dimos}).
Florian would like to thank Christina Isaicu for her advice and continued support throughout this project.
}


\bibliography{main}  

\begin{thebibliography}{21}
\providecommand{\natexlab}[1]{#1}
\providecommand{\url}[1]{\texttt{#1}}
\expandafter\ifx\csname urlstyle\endcsname\relax
  \providecommand{\doi}[1]{doi: #1}\else
  \providecommand{\doi}{doi: \begingroup \urlstyle{rm}\Url}\fi

\bibitem[Bourne et~al.(2021)Bourne, Steinmetz, Flaxman, et~al.]{bourne_trends_2021}
R.~R. Bourne, J.~D. Steinmetz, S.~Flaxman, et~al.
\newblock Trends in prevalence of blindness and distance and near vision impairment over 30 years: an analysis for the global burden of disease study.
\newblock \emph{The Lancet Global Health}, 9\penalty0 (2):\penalty0 e130--e143, 2021.
\newblock ISSN 2214-109X.
\newblock \doi{10.1016/S2214-109X(20)30425-3}.
\newblock URL \url{https://www.thelancet.com/journals/langlo/article/PIIS2214-109X(20)30425-3/fulltext}.

\bibitem[{CNIB Foundation}(2025)]{guide_dog_foundation}
{CNIB Foundation}.
\newblock Guide the way, 2025.
\newblock URL \url{https://www.guidedog.org/gd/about-us/about-the-guide-dog-foundation.aspx}.
\newblock Accessed: 2026-05-28.

\bibitem[Sorokin et~al.(2022)Sorokin, Tan, Liu, and Ha]{sorokin2022learning}
M.~Sorokin, J.~Tan, C.~K. Liu, and S.~Ha.
\newblock Learning to navigate sidewalks in outdoor environments.
\newblock \emph{IEEE Robotics and Automation Letters}, 7\penalty0 (2):\penalty0 3906--3913, 2022.

\bibitem[Cai et~al.(2024)Cai, Ram, Gou, Shaikh, Chen, Wan, Hara, Zhao, and Hsu]{cai2024navigating}
S.~Cai, A.~Ram, Z.~Gou, M.~A.~W. Shaikh, Y.-A. Chen, Y.~Wan, K.~Hara, S.~Zhao, and D.~Hsu.
\newblock Navigating real-world challenges: A quadruped robot guiding system for visually impaired people in diverse environments.
\newblock In \emph{Proceedings of the 2024 CHI Conference on Human Factors in Computing Systems}, pages 1--18, 2024.

\bibitem[Hwang et~al.(2022)Hwang, Xia, Keita, Suzuki, Biswas, Lee, and Kim]{hwang2022system}
H.~Hwang, T.~Xia, I.~Keita, K.~Suzuki, J.~Biswas, S.~I. Lee, and D.~Kim.
\newblock System configuration and navigation of a guide dog robot: Toward animal guide dog-level guiding work.
\newblock \emph{arXiv preprint arXiv:2210.13368}, 2022.

\bibitem[Hwang et~al.(2025)Hwang, Yang, Monon, Giudice, Lee, Biswas, and Kim]{hwang_guidenav_2025}
H.~Hwang, S.~Yang, J.~S. Monon, N.~A. Giudice, S.~I. Lee, J.~Biswas, and D.~Kim.
\newblock {GuideNav}: User-informed development of a vision-only robotic navigation assistant for blind travelers, 2025.
\newblock URL \url{http://arxiv.org/abs/2512.06147}.

\bibitem[Hayamizu et~al.(2026)Hayamizu, {DeFazio}, Mehta, Altaweel, Choe, Lin, Juettner, Xiao, Blackburn, and Zhang]{hayamizu_woofs_2026}
Y.~Hayamizu, D.~{DeFazio}, H.~Mehta, Z.~Altaweel, J.~Choe, C.~Lin, J.~Juettner, F.~Xiao, J.~Blackburn, and S.~Zhang.
\newblock From woofs to words: Towards intelligent robotic guide dogs with verbal communication, 2026.
\newblock URL \url{http://arxiv.org/abs/2603.12574}.

\bibitem[Hwang et~al.(2024)Hwang, Jung, Giudice, Biswas, Lee, and Kim]{hwang_towards_2024}
H.~Hwang, H.-T. Jung, N.~A. Giudice, J.~Biswas, S.~I. Lee, and D.~Kim.
\newblock Towards robotic companions: Understanding handler-guide dog interactions for informed guide dog robot design, 2024.
\newblock URL \url{http://arxiv.org/abs/2402.06790}.

\bibitem[Hu et~al.(2019)Hu, Li, Anderson, Ragan-Kelley, and Durand]{hu2019taichi}
Y.~Hu, T.-M. Li, L.~Anderson, J.~Ragan-Kelley, and F.~Durand.
\newblock Taichi: a language for high-performance computation on spatially sparse data structures.
\newblock \emph{ACM Transactions on Graphics (TOG)}, 38\penalty0 (6):\penalty0 201, 2019.

\bibitem[Chen and Zhang(2025)]{chen_exploration_2025}
J.~Chen and B.~Zhang.
\newblock Exploration and navigation in unknown environments for guide dog robots.
\newblock In \emph{2025 9th International Conference on Robotics and Automation Sciences ({ICRAS})}, pages 48--326, 2025.
\newblock \doi{10.1109/ICRAS65818.2025.11108808}.
\newblock URL \url{https://ieeexplore.ieee.org/document/11108808}.
\newblock {ISSN}: 2694-3506.

\bibitem[Viteri and Li(2024)]{viteri_autonomous_2024}
J.~Viteri and C.-H.~G. Li.
\newblock Autonomous sidewalk navigation featuring end-to-end {RGB}-d dual-{ConvNet} steering.
\newblock In \emph{2024 {IEEE} International Conference on Advanced Intelligent Mechatronics ({AIM})}, pages 703--708. {IEEE}, 2024.
\newblock ISBN 979-8-3503-5536-9.
\newblock \doi{10.1109/AIM55361.2024.10637141}.
\newblock URL \url{https://ieeexplore.ieee.org/document/10637141/}.

\bibitem[Kim et~al.(2023)Kim, Yu, Kothari, Tan, Turk, and Ha]{kim2023transforming}
J.~T. Kim, W.~Yu, Y.~Kothari, J.~Tan, G.~Turk, and S.~Ha.
\newblock Transforming a quadruped into a guide robot for the visually impaired: Formalizing wayfinding, interaction modeling, and safety mechanism.
\newblock \emph{arXiv preprint arXiv:2306.14055}, 2023.

\bibitem[Zhang et~al.(2023)Zhang, Han, Qiao, Kim, Bae, Lee, and Hong]{mobile_sam}
C.~Zhang, D.~Han, Y.~Qiao, J.~U. Kim, S.-H. Bae, S.~Lee, and C.~S. Hong.
\newblock Faster segment anything: Towards lightweight sam for mobile applications.
\newblock \emph{arXiv preprint arXiv:2306.14289}, 2023.

\bibitem[{NVIDIA}(2024)]{nvidia_nanosam}
{NVIDIA}.
\newblock Nanosam.
\newblock \url{https://www.jetson-ai-lab.com/archive/vit/tutorial_nanosam.html}, 2024.
\newblock Accessed: 2026-05-28.

\bibitem[Jocher and Qiu(2026)]{yolo26_ultralytics}
G.~Jocher and J.~Qiu.
\newblock Ultralytics yolo26, 2026.
\newblock URL \url{https://github.com/ultralytics/ultralytics}.

\bibitem[Yang et~al.(2024)Yang, Kang, Huang, Zhao, Xu, Feng, and Zhao]{depth_anything_v2}
L.~Yang, B.~Kang, Z.~Huang, Z.~Zhao, X.~Xu, J.~Feng, and H.~Zhao.
\newblock Depth anything v2.
\newblock \emph{arXiv:2406.09414}, 2024.

\bibitem[Rudin et~al.(2021)Rudin, Hoeller, Reist, and Hutter]{rudin2021learning}
N.~Rudin, D.~Hoeller, P.~Reist, and M.~Hutter.
\newblock Learning to walk in minutes using massively parallel deep reinforcement learning. arxiv.
\newblock \emph{arXiv preprint arXiv:2109.11978}, 2021.

\bibitem[Liang et~al.(2022)Liang, Xie, Yu, Xia, Lin, Wang, Tang, Wang, and Tang]{liang2022bevfusion}
T.~Liang, H.~Xie, K.~Yu, Z.~Xia, Z.~Lin, Y.~Wang, T.~Tang, B.~Wang, and Z.~Tang.
\newblock Bevfusion: A simple and robust lidar-camera fusion framework.
\newblock \emph{Advances in Neural Information Processing Systems}, 35:\penalty0 10421--10434, 2022.

\bibitem[Xie et~al.(2021)Xie, Wang, Yu, Anandkumar, Alvarez, and Luo]{DBLP:journals/corr/abs-2105-15203}
E.~Xie, W.~Wang, Z.~Yu, A.~Anandkumar, J.~M. Alvarez, and P.~Luo.
\newblock Segformer: Simple and efficient design for semantic segmentation with transformers.
\newblock \emph{CoRR}, abs/2105.15203, 2021.
\newblock URL \url{https://arxiv.org/abs/2105.15203}.

\bibitem[Prautzsch et~al.(2002)Prautzsch, Boehm, and Paluszny]{prautzsch2002bezier}
H.~Prautzsch, W.~Boehm, and M.~Paluszny.
\newblock \emph{B{\'e}zier and B-spline techniques}, volume~6.
\newblock Springer, 2002.

\bibitem[Perlin(1985)]{perlin1985improved}
K.~Perlin.
\newblock An image synthesizer.
\newblock \emph{ACM SIGGRAPH Computer Graphics}, 19\penalty0 (3):\penalty0 287--296, 1985.
\newblock \doi{10.1145/325165.325247}.

\end{thebibliography}
\newpage
\begin{appendices}



\section{Evaluation of Walkable Surface Segmentation Methods}

\begin{figure}[h]
\centering
\includegraphics[width=1\textwidth]{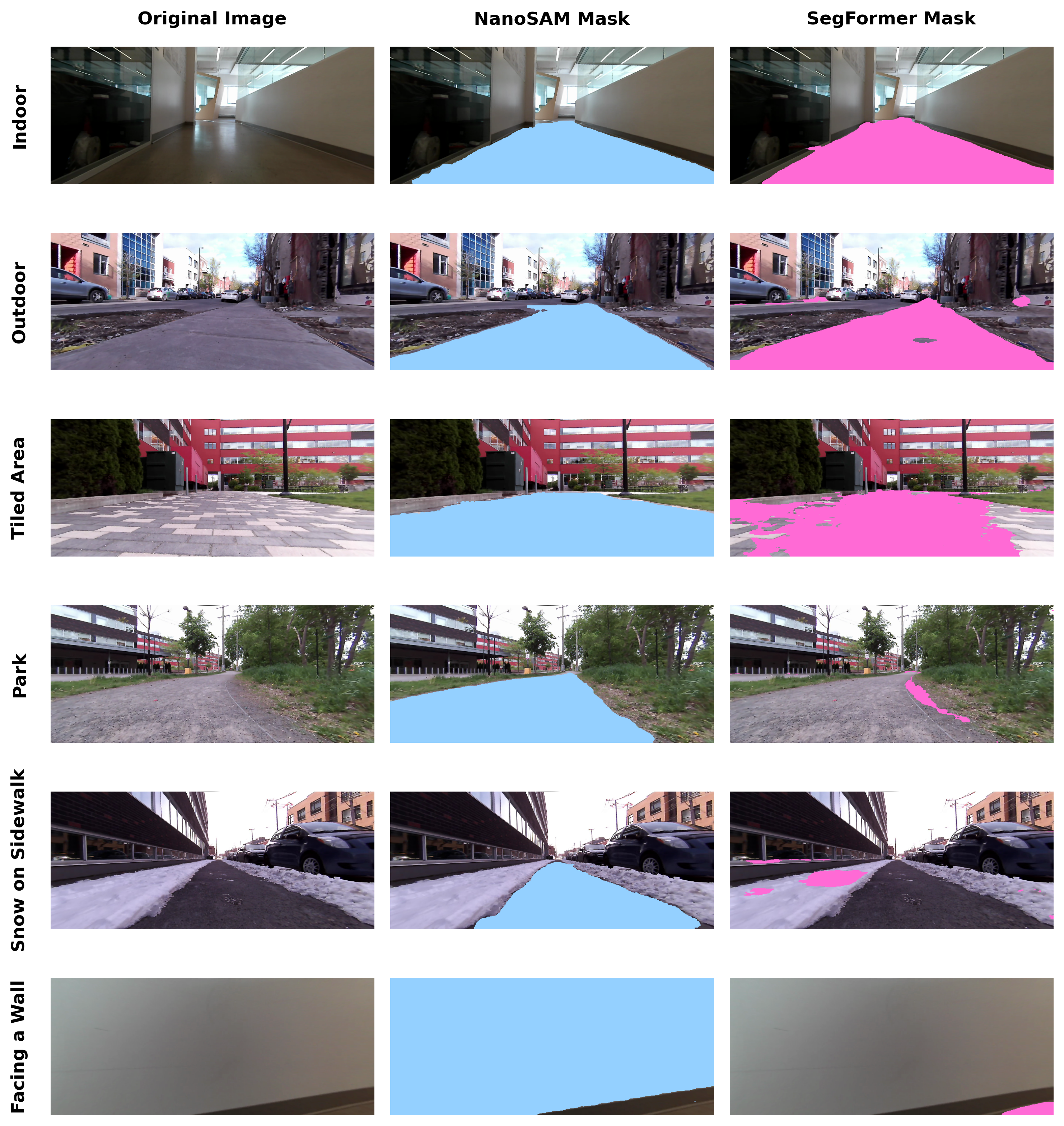}
\caption{\textbf{Comparison of NanoSAM and SegFormer walkable-surface segmentation.} Columns show the input RGB image, NanoSAM prediction, and SegFormer prediction, respectively. Rows correspond to an indoor floor, an outdoor sidewalk, a tiled pedestrian area, a park path, a sidewalk partially covered by snow, and a camera frame where the robot is facing a wall. Both methods perform well on standard indoor-floor and outdoor-sidewalk examples. NanoSAM produces more consistent segmentations on tiled surfaces and better generalizes to park paths and snow-covered sidewalks. However, when the robot is facing a wall, NanoSAM incorrectly segments the wall as walkable because of its prompt-based formulation, whereas SegFormer successfully identifies the small visible floor region. These examples highlight the trade-off between the stronger generalization capabilities of NanoSAM and the orientation robustness of SegFormer.
}
\label{fig:comparison_segmentation}
\end{figure}

As an alternative to NanoSAM, we evaluated walkable-surface segmentation approaches based on SegFormer \cite{DBLP:journals/corr/abs-2105-15203}, including models fine-tuned for outdoor sidewalk segmentation\footnote{\url{https://huggingface.co/tobiasc/segformer-b0-finetuned-segments-sidewalk}} and indoor floor segmentation\footnote{\url{https://huggingface.co/nvidia/segformer-b0-finetuned-ade-512-512}}. 
These models directly predict dense segmentation masks from RGB images, enabling the identification of walkable areas without requiring user-defined prompt points. 
Unlike NanoSAM, SegFormer does not rely on assumptions about the robot's initial orientation relative to the path, making it more robust in situations where the robot is not facing a traversable surface, as illustrated by the last row of Fig.~\ref{fig:comparison_segmentation}.

However, separate models are required for indoor-floor and outdoor-sidewalk segmentation, necessitating model switching when transitioning between indoor and outdoor environments. 
In addition, we observed degraded performance in scenarios that differ from the training distribution, such as snow-covered sidewalks, tiled surfaces, and park paths.
Fig.~\ref{fig:comparison_segmentation} highlights the complementary strengths and limitations of the two approaches.
We opted for SAM/NanoSAM because the main downside is easy to address: starting the robot not directly facing a wall.


\section{Pedestrian Depth Estimation}
\label{app:depth}

The LiDAR-based obstacle map used by the navigation system is updated at a relatively low frequency and covers only a local region around the robot. 
While sufficient for static obstacle avoidance, this representation can introduce latency when tracking moving pedestrians. To improve responsiveness, we evaluated two camera-based pedestrian localization methods.

The first method uses YOLO pedestrian detections and projects the center of the lower edge of each bounding box onto the ground plane using a precomputed camera calibration. This approach is lightweight and suitable for onboard deployment, but the resulting position estimates are often noisy due to frame-to-frame variations in the bounding box location.

The second method combines YOLO detections with monocular depth estimation using a fine-tuned version of Depth Anything V2 for indoor metric depth estimation\footnote{\url{https://huggingface.co/depth-anything/Depth-Anything-V2-Metric-Indoor-Small-hf}} \cite{depth_anything_v2}.
Although the model directly predicts metric depth, its estimates are further calibrated using LiDAR measurements to improve consistency with the robot's sensing setup. 
The pedestrian distance is estimated from the closest 10\% of depth values within each detected bounding box. 
Although the model is fine-tuned for indoor scenes, preliminary testing showed that it also produced reliable depth estimates in our outdoor environment over the distance range considered in this work.
Compared with the floor-projection approach, this method produces smoother and more accurate distance estimates, particularly at longer ranges.

To enable real-time deployment on the Jetson Orin, we adopted the TensorRT optimization pipeline proposed by IRCVLab\footnote{\url{https://github.com/IRCVLab/Depth-Anything-for-Jetson-Orin}} while replacing the original model weights with those of the fine-tuned Depth Anything V2 model for indoor metric depth estimation.
This optimized implementation provides distance estimates that are comparable in accuracy to the fine-tuned metric Depth Anything V2 implementation while increasing the inference rate from 2.64 FPS to 24.02 FPS, making it suitable for real-time onboard deployment on \milo.

\begin{figure}[htb]
\centering
\includegraphics[width=1\textwidth]{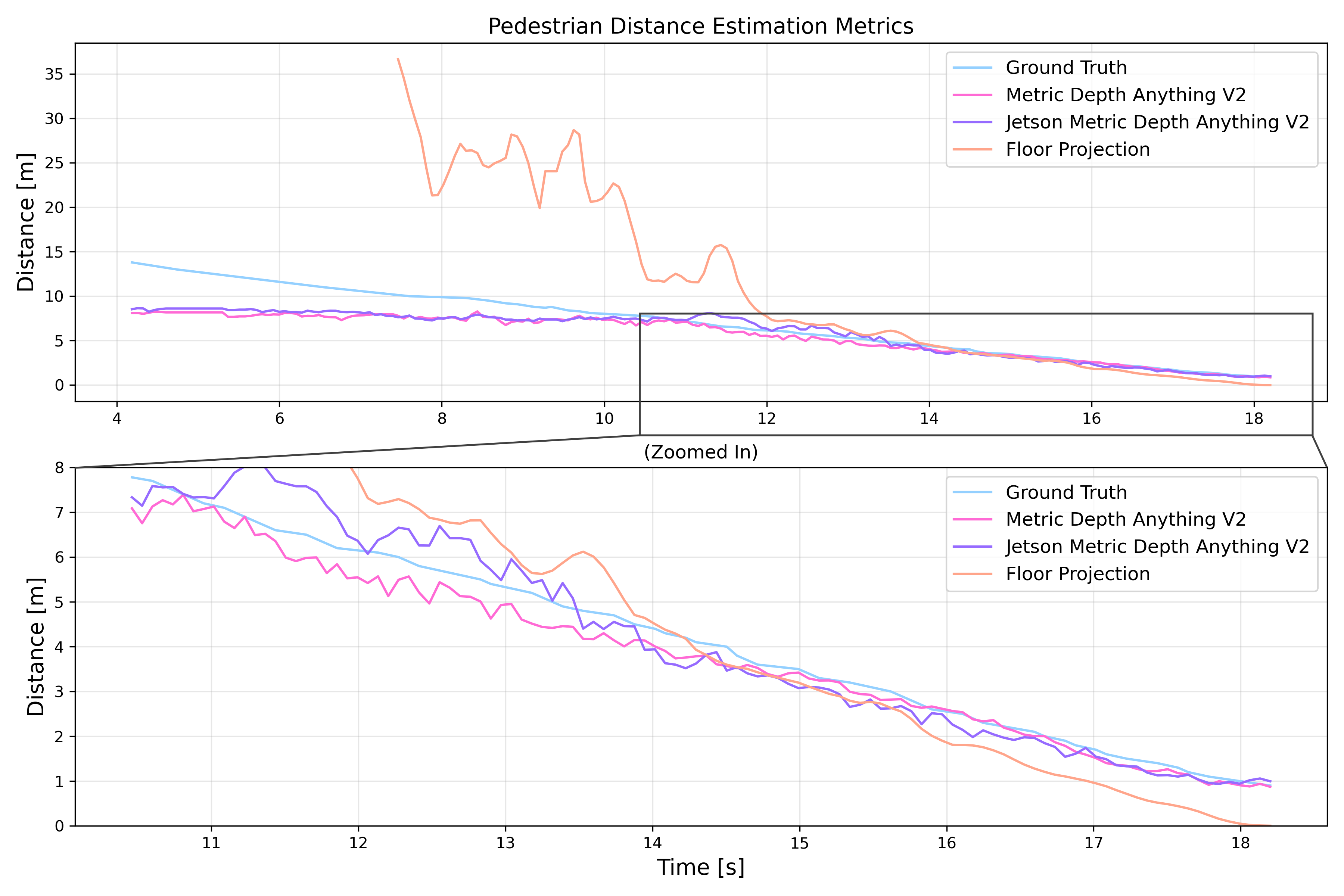}
\caption{\textbf{Estimated distance to a pedestrian as a function of time for several camera-based localization approaches.} All methods were evaluated on the same video recording of a person walking toward the robot. Ground-truth distances were obtained using an AirTag carried by the pedestrian and were interpolated to produce a smooth reference trajectory. The figure compares three methods for estimating the distance of pedestrians detected by YOLO: a floor-projection approach that projects the center of the lower edge of each YOLO bounding box onto the ground plane using the camera calibration, a fine-tuned version of Depth Anything V2 for indoor metric depth estimation calibrated using LiDAR measurements, and a Jetson-optimized implementation based on the IRCVLab TensorRT pipeline using the same fine-tuned model weights and LiDAR calibration. The floor-projection method exhibits significant noise at longer ranges due to frame-to-frame variations in bounding box localization, whereas depth-based methods provide smoother and more accurate estimates. The two Metric Depth Anything V2 implementations achieve comparable accuracy across the evaluated range, while the Jetson-optimized version provides substantially higher inference speed, enabling real-time deployment on \milo.
}
\label{fig:person_distance}
\end{figure}

Fig.~\ref{fig:person_distance} compares the different methods for estimating the distance between a pedestrian and the robot. Ground-truth measurements were obtained using an AirTag carried by a person walking toward the robot, and all methods were evaluated on the same video recording.
The floor-projection approach exhibits considerable noise, particularly at longer distances, due to frame-to-frame variations in the detected bounding boxes.
Both implementations of the fine-tuned Depth Anything V2 model for indoor metric depth estimation produce smoother and more accurate distance estimates.
Their performance is similar over the evaluated range, while the TensorRT-optimized Jetson implementation achieves a significantly higher inference rate (approximately 24.02 FPS compared with 2.64 FPS), making it suitable for real-time onboard deployment on \milo.

\section{Local Obstacle Safety Filter}
\label{app:safety}

The Local Obstacle Safety Filter is a reactive safety layer that filters the velocity commands before they are executed by the robot. 
Its primary purpose is to prevent collisions with nearby obstacles, protecting the robot from damage regardless of whether navigation is performed using the costmap-based planner or the learned navigation policy. 
This additional safety layer is necessary because the costmap does not observe obstacles located in the camera blind spots along the sides of the robot, while the learned policy may intentionally choose trajectories that bring the robot close to obstacles in order to maximize the clearance around the handler, whose safety is prioritized in the navigation objective.

The filter operates directly on the raw LiDAR point cloud together with the robot pose and only modifies velocity components that would move the robot too close to an obstacle. 
As the robot approaches an obstacle in front, the commanded forward velocity is progressively reduced and eventually set to zero if a critical safety distance is reached. 
Similarly, lateral motions directed toward nearby obstacles are first blocked, and if the robot is already within a critical distance, a small lateral velocity is applied in the opposite direction to increase the clearance.

Tab.\ref{tab:safety_filter} illustrates examples of the safety filter in simulation. 
The rectangle represents the robot. 
The green arrow indicates the filtered velocity command sent to the robot, while the red arrow denotes the original desired velocity generated by the navigation policy whenever it is modified by the safety filter. 
The forward and lateral velocity components correspond to the robot's local x- and y-axes, respectively.




\begin{table*}[t]
\centering
\caption{Examples of the Local Obstacle Safety Filter behavior.}
\label{tab:safety_filter}

\renewcommand{\arraystretch}{1.2}
\setlength{\tabcolsep}{3pt}

\begin{tabular}{ll|ccc}
\toprule
& & \multicolumn{3}{c}{\textbf{Desired motion}} \\
\cmidrule(lr){3-5}
\textbf{Wall Placement} & \textbf{Distance} & \textbf{Forward Velocity} & \textbf{Lateral Velocity} & \textbf{Diagonal Velocity} \\
\midrule

\multirow{3}{*}{Facing Wall}
& Safe &
\includegraphics[width=0.18\linewidth]{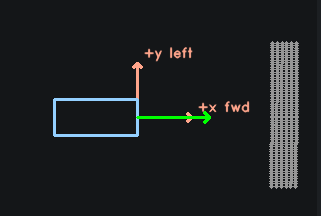} &
\includegraphics[width=0.18\linewidth]{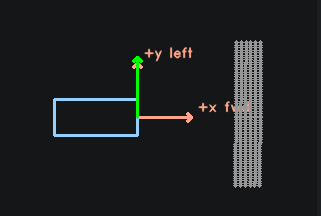} &
\includegraphics[width=0.18\linewidth]{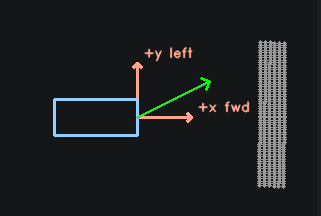} \\

& Near &
\includegraphics[width=0.18\linewidth]{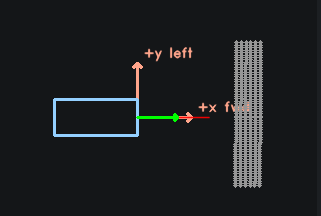} &
\includegraphics[width=0.18\linewidth]{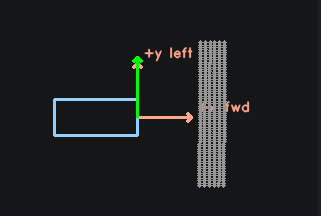} &
\includegraphics[width=0.18\linewidth]{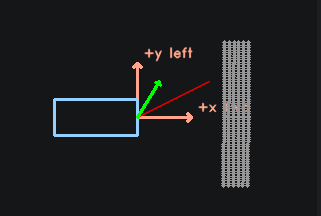} \\

& Critical &
\includegraphics[width=0.18\linewidth]{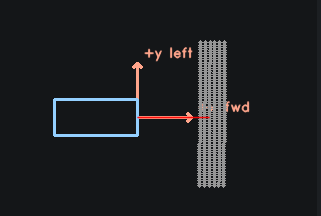} &
\includegraphics[width=0.18\linewidth]{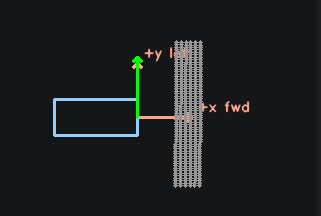} &
\includegraphics[width=0.18\linewidth]{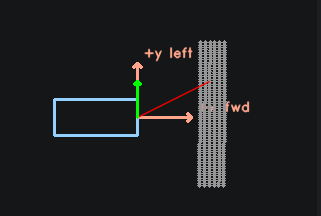} \\

\midrule

\multirow{3}{*}{Side Wall}
& Safe &
\includegraphics[width=0.18\linewidth]{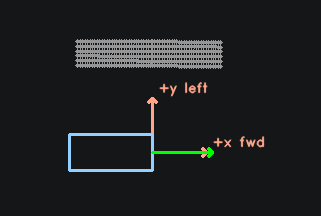} &
\includegraphics[width=0.18\linewidth]{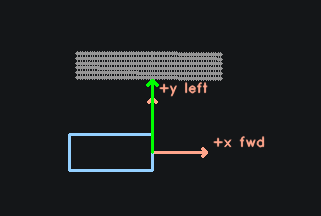} &
\includegraphics[width=0.18\linewidth]{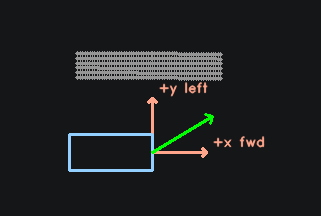} \\

& Near &
\includegraphics[width=0.18\linewidth]{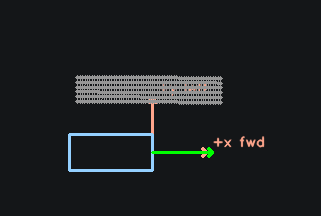} &
\includegraphics[width=0.18\linewidth]{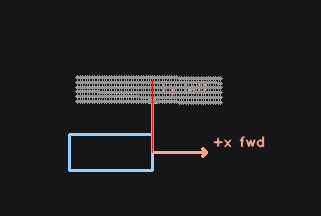} &
\includegraphics[width=0.18\linewidth]{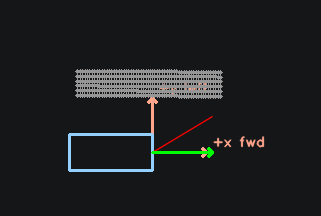} \\

& Critical &
\includegraphics[width=0.18\linewidth]{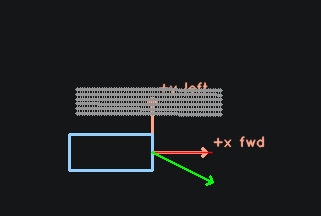} &
\includegraphics[width=0.18\linewidth]{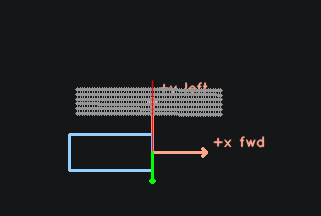} &
\includegraphics[width=0.18\linewidth]{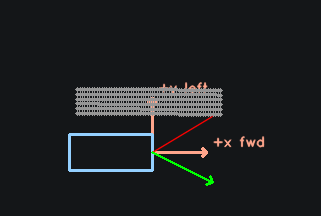} \\

\midrule

\multirow{3}{*}{Diagonal Wall}
& Safe &
\includegraphics[width=0.18\linewidth]{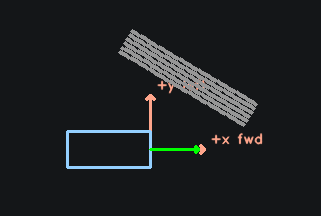} &
\includegraphics[width=0.18\linewidth]{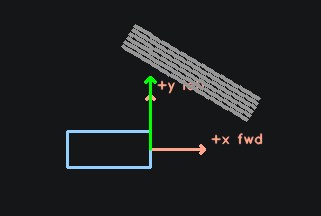} &
\includegraphics[width=0.18\linewidth]{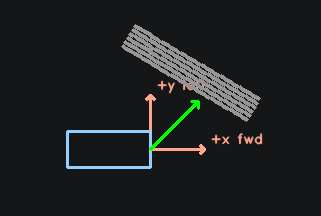} \\

& Near &
\includegraphics[width=0.18\linewidth]{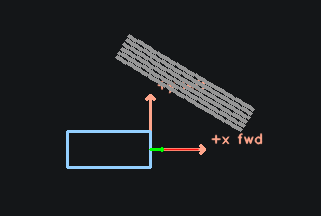} &
\includegraphics[width=0.18\linewidth]{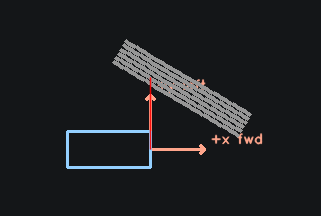} &
\includegraphics[width=0.18\linewidth]{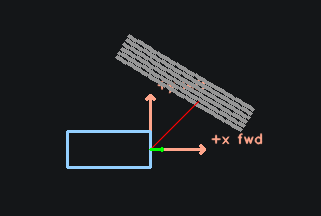} \\

& Critical &
\includegraphics[width=0.18\linewidth]{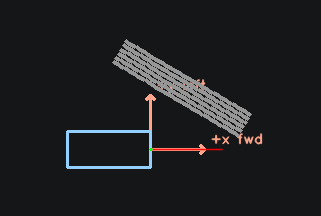} &
\includegraphics[width=0.18\linewidth]{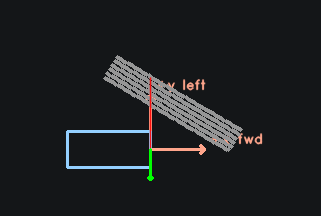} &
\includegraphics[width=0.18\linewidth]{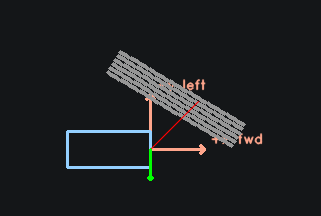} \\

\bottomrule
\end{tabular}
\end{table*}

\section{Simulator Implementation}
\label{app:sim}

\begin{figure}[h]
  \centering
  \includegraphics[width=0.5\textwidth]{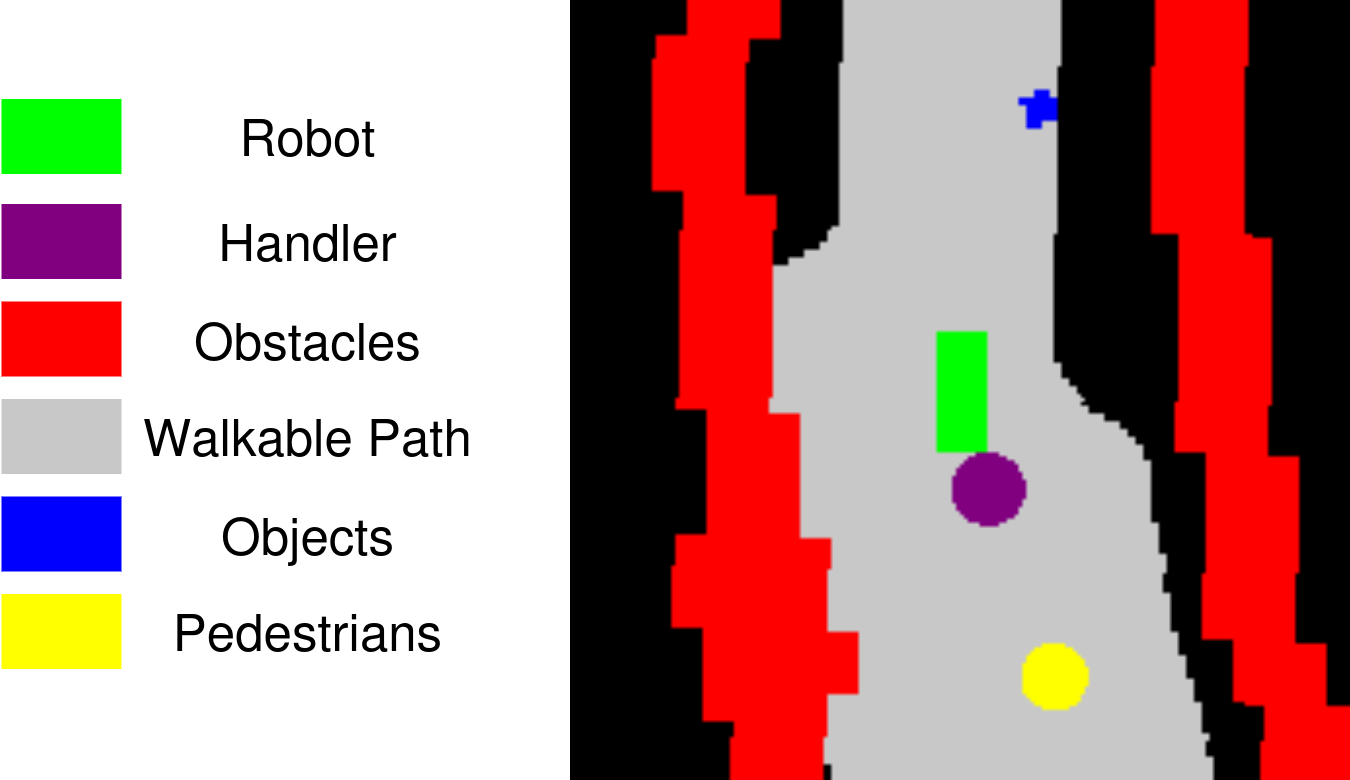}
  \caption{\textbf{Example Simulator Frame.} Visualization of the simulated environment. The traversable road is shown in grey (with intentional gaps to mimic imperfect floor segmentation), walls in red, handler in purple, pedestrian in yellow, and obstacle in blue.
  }
  \label{fig:frame-sim}
  \end{figure}

Our birds-eye-view simulator, implemented in Taichi \cite{hu2019taichi}, constructs scenes like this:
Roads are random lines and Bezier curves \cite{prautzsch2002bezier} drawn in OpenCV that can be rotated and combined for tee and cross intersections.
Roads are randomly redrawn whenever the episode resets (after 100 steps in the environment).
The size of the roads is sampled in the range of sidewalk widths allowed in the civil code of our city.
2D Perlin noise \cite{perlin1985improved} is applied to the road map to approximate incomplete floor segmentation. 
The walls are road contours that are randomly sampled (to allow for segments with no walls, walls only on either the left or right side, or walls on both sides) and drawn on top of the roads.
The road lines/curves are made up of points, which serve as spawn points for the robot.
Before the robot is placed, a small random offset and random rotation are added, but the robot is spawned generally spawned towards the edge of the map, facing inward.
The camera is always tracking the center of the robot with the robot facing upward (i.e., the rest of the map is rotated, zoomed, and cropped to keep a 8m or 4m local area around the robot in view\footnote{Ideally, we would have wanted the larger 8m viewport for smoother obstacle and pedestrian avoidance, but we found empirically that the L2 LiDAR on the Go2 is not reliable enough for this distance. In practice, we used the 4m view port for all experiments. We leave better integration of depth estimation networks to improve the LiDAR accuracy not just for pedestrians but also for the path geometry to future work.}).
On top of the scene, we draw the handler (a circle at a random position parameterized by azimuth, elevation, and length of the handle), obstacles (randomly rotated boxes, with positions sampled between the center line and the edge of the road), and pedestrians (circles, following the path segments at a fixed left/right offset and random fixed speed).
On an Nvidia Titan RTX GPU from 2018, this environment can be instantiated 64 times and runs each instance at roughly 40 frames per second, leading to a total throughput of approximately 2560 frames per second, each rendered at 200x200 pixels. 


\section{Reinforcement Learning}
\label{app:rl}

We use the following reward terms:
The binary collision penalties for the dog or handler colliding with the walls is $r_{coll\_wall}$, obstacles $r_{coll\_obst}$, and pedestrians $r_{coll\_ped}$ and for going off-road $r_{coll\_off}$.

The term for maintaining the default forward velocity is 
\begin{equation}
    r_{vel} = \exp(-\frac{(v_{fwd\_act} - v_{fwd\_target})^2}{\sigma^2})
\end{equation}
where $v_{fwd\_act}$ is the forward velocity component of the policy's action output, $v_{fwd\_target}$ is the target velocity (for us, set to 0.9 m/s, determined empirically by walking speed comfort) and $\sigma$ the smoothness of the reward function (for us, set to 0.3, found empirically).

The terms for keeping the lateral movement and yaw small are
\begin{align}
    r_{lat} &= (\frac{v_{lat\_act}}{v_{lat\_max}})^2\\
    r_{yaw} &= (\frac{v_{yaw\_act}}{v_{yaw\_max}})^2
\end{align}
where $v_{lat\_act}$ and $v_{yaw\_act}$ are the lateral movement and yaw components of the policy's action output, and $v_{lat\_max}$ and $v_{yaw\_max}$ are the maximum lateral and yaw velocities (for us, 1.5 m/s and 0.5 rad/s, determined by the Go2 robot)

The terms for measuring obstacles in the safety zones around the robot and handler are:
\begin{align}
r_{inner\_danger} &= r_{occup}(x, y, R_{inner})\\
r_{outer\_danger} &= 1 - r_{occup}(x, y, R_{outer})
\end{align}
where $r_{occup}(x, y, R)$ is the function that measures the number of obstacle pixels inside the radius around the point, normalized by the total number of pixels in the circle and is defined as
\begin{equation}
    r_{occup}(x, y, R) = \frac{1}{N_{\max}} \sum_{(u,v) \in \mathcal{D}(x,y,R)} \mathbf{1}\bigl[\text{pixel}_{u,v} \text{ belongs to wall/obst/pedestr.}\bigr]
\end{equation}
where $N_{\max}$ is the maximum number of pixels in the circle and $\mathcal{D}(x,y,R)$ is the circular set function that contains all points $(u,v)$ within a radius $R$ around a mid point $(x,y)$ defined as
\begin{equation}
\mathcal{D}(x,y,R)
=
\left\{(u,v) \;|\; (u - x)^2 + (v - y)^2 \le R^2\right\}
\end{equation}
We use 1.2m as the outer radius and 0.6m as the inner radius.
These were chosen empirically, with smaller values leading to more aggressive and dangerous policies and larger values leading to more ``skittish'' policies.
The center of the safety circle is the midpoint between the center of the dog and the center of the handler's position.

The jerk penalty is expressed as the squared difference between the last action $a_{t-1}$ and the current action $a_t$.
\begin{equation}
    r_{jerk}=||a_{t-1} - a_{t}||^2 
\end{equation}


The full reward consists of the following reward terms and their coefficients:
\begin{table}[h]
\centering
\begin{tabular}{lrl}
\hline
Reward term & Coefficient & Description \\
\hline
$r_{coll\_wall}$        & $-4.0$   & Collisions between dog/handler and wall \\
$r_{coll\_ped}$         & $-3.0$   & Collisions between dog/handler and pedestrians \\
$r_{coll\_obst}$        & $-2.0$   & Collisions between dog/handler and obstacles \\
$r_{coll\_off}$         & $-0.1$ & Dog/handler leaving the visible road \\
$r_{vel}$          & $2.0$    & Tracking the forward velocity \\
$r_{lat}$        & $-0.1$  & Keeping lateral velocity small \\
$r_{yaw}$        & $-0.1$  & Keeping yaw rate small \\
$r_{jerk}$              & $-0.1$ & Difference between previous and current action \\
$r_{inner\_danger}$     & $-2.0$   & Obstacles/walls/pedestrians inside inner safety circle \\
$r_{outer\_danger}$     & $1.0$    & Obstacles/walls/pedestrians inside outer safety circle \\
\hline
\end{tabular}
\caption{Reward terms, coefficients, and descriptions.}
\label{tab:reward_terms}
\end{table}

These reward coefficients have been determined empirically based on success rate in the simulated environment, defined by maintaining forward velocity while minimizing collisions.

The hyperparameters for PPO can be found in Tab.\ref{tab:ppo}. 
Except for the number of iterations, all of these are the default values of the RSL-RL library from \citet{rudin2021learning}.
The number of iterations is slightly more than what is required for cumulative reward to plateau in our environment.




\begin{table}[htb]
\centering
\caption{PPO and CNN Hyperparameters}
\begin{tabular}{ll}
\hline
\textbf{Parameter} & \textbf{Value} \\
\hline
\multicolumn{2}{c}{\textit{PPO Configuration}} \\
\hline
Training iterations & 2000 \\
Steps per environment & 64 \\
Parallel environments & 10 \\
Clip parameter & 0.2 \\
Learning epochs & 5 \\
Mini-batches & 4 \\
Learning rate & \(3 \times 10^{-4}\) \\
Gamma & 0.99 \\
Lambda (GAE) & 0.95 \\
Entropy coefficient & 0.01 \\
Desired KL & 0.01 \\
Max grad norm & 1.0 \\
\hline
\multicolumn{2}{c}{\textit{Network Configuration}} \\
\hline
Observation normalization & True \\
Output channels & [32, 64, 64] \\
Kernel sizes & [8, 4, 3] \\
Strides & [4, 2, 1] \\
Activation & ELU \\
\hline
\label{tab:ppo}
\end{tabular}
\end{table}




\end{appendices}
\end{document}